\documentclass[runningheads]{llncs}
\usepackage[T1]{fontenc}
\usepackage{graphicx}
\usepackage{hyperref}
\usepackage{color}

\usepackage{tabularx}
\usepackage[dvipsnames,x11names]{xcolor}
\graphicspath{{figs/}}
\usepackage{amssymb}
\begin{document}
\title{Tract-RLFormer: A Tract-Specific RL policy based Decoder-only Transformer Network}
\titlerunning{Tract-RLFormer}
%
\author{Ankita Joshi\inst{1} \and
Ashutosh Sharma\inst{1} \and
Anoushkrit Goel\inst{1} \and
Ranjeet Ranjan Jha\inst{2} \and
Chirag Ahuja\inst{3} \and
Arnav Bhavsar\inst{1} \and
Aditya Nigam\inst{1}}
\authorrunning{A. Joshi et al.}
%
\institute{Indian Institute of Technology (IIT) Mandi, India \and
Indian Institute of Technology (IIT) Patna, India \and
Post-Graduate Inst. of Medical Edu. and Research (PGIMER), Chandigarh, India}
\maketitle              
\begin{abstract}
Fiber tractography is a cornerstone of neuroimaging, enabling the detailed mapping of the brain's white matter pathways through diffusion MRI. This is crucial for understanding brain connectivity and function, making it a valuable tool in neurological applications. Despite its importance, tractography faces challenges due to its complexity and susceptibility to false positives, misrepresenting vital pathways. To address these issues, recent strategies have shifted towards deep learning, utilizing supervised learning, which depends on precise ground truth, or reinforcement learning, which operates without it. In this work, we propose Tract-RLFormer, a network utilizing both supervised and reinforcement learning, in a two-stage policy refinement process that markedly improves the accuracy and generalizability across various data-sets. By employing a tract-specific approach, our network directly delineates the tracts of interest, bypassing the traditional segmentation process. Through rigorous validation on datasets such as TractoInferno, HCP, and ISMRM-2015, our methodology demonstrates a leap forward in tractography, showcasing its ability to accurately map the brain's white matter tracts.

\end{abstract}

\section{Introduction}

Tractography is an advanced reconstruction technique in neuroscience, that leverages diffusion MRI to create detailed visual representations of brain's white matter pathways.
This technology has played a crucial role in assisting neurosurgeons with meticulous pre-surgical planning, benefiting patients with a range of neurological disorders \cite{essayed2017white}, by enabling a deeper analysis of the white matter. Over the years, a range of tractography algorithms have been developed, to map critical neurological pathways. Deterministic algorithms \cite{basser1998fiber} trace fiber paths directly based on the most probable direction of water molecule diffusion, offering clear but sometimes oversimplified views of white matter tracts. In contrast, probabilistic algorithms \cite{berman2008probabilistic} incorporate the inherent uncertainty in diffusion data to predict multiple potential pathways, resulting in detailed fiber reconstruction. Global algorithms \cite{fillard2009novel} attempt to reconcile the deterministic and probabilistic approaches by optimizing whole-brain tractography reconstructions to capture the complex architecture of brain connectivity.

Despite these advancements, tractography still faces challenges such as the \textit{crossing-fibers} issue (also known as the \textit{bottleneck} phenomenon) \cite{maier2017challenge} due to its ill-posed nature. These issues arise because the algorithms rely on local diffusion information to reconstruct the brain's complete fiber network, occasionally resulting in erroneous projection of fiber pathways or false positive connections.

To overcome these obstacles, recent research employs machine learning and deep learning (DL) approaches to enhance tractography accuracy.
Supervised DL techniques \cite{poulin2017learn,benou2019deeptract} for tractography rely on accurate and comprehensive ground truth data to train and validate the algorithms, which is very difficult to obtain. In this regard, recent works \cite{theberge2021track,theberge2024matters} have proposed deep reinforcement learning (DRL)-based approaches, that learn to perform tractography by interacting with the environment.
These techniques, leveraging deep neural networks, enhance the ability to predict brain fiber configurations, promising significant advancements in fiber mapping quality for neurological research and clinical applications. However, improving tractography algorithms for effective use across diverse datasets remains a challenge for further research in the field.

Recently, transformers have shown remarkable performance in various domains, including language modeling\cite{vaswani2017attention}, image recognition\cite{dosovitskiy2020image}, time series fore\-casting\cite{zhou2021informer}, and even protein structure prediction\cite{jumper2021highly}. Their robust performance in various sequence prediction tasks demonstrates their ability to capture long-range dependencies and contextual information effectively, making them well-suited for mapping neural pathways in tractography. Building on their success in related fields, we now extend the generalization, transfer learning, and autoregressive capabilities of transformers (GPT), to the tractography domain \textbf{in a novel hybrid framework}.
We adopt an RL framework (\cite{chen2021decision}) to generate training data for our GPT model, Tract-RLFormer, reducing the need for extensive ground-truth typically needed to train transformers. This addresses a significant challenge of applying transformers for tractography where ground-truth fibers are very difficult to obtain.

This approach also represents a significant departure from traditional methods, as it simplifies the tractography process by targeting specific tracts, thereby eliminating the need for complex and often cumbersome segmentation algorithms employed post-tractography. Moreover our data-driven approach has the potential to utilize data from RL agents trained across diverse neuroimaging environments.
Our key contributions are as follows:\\
\begin{enumerate}
\vspace{-0.7cm}
\item \textbf{Data driven policy learning via hybrid framework}: 
We propose Tract-RLFormer, a GPT-based network trained by leveraging both reinforcement learning (RL) and supervised learning (SL) paradigms, to approximate and refine a policy for tract generation that outperforms recent RL-algorithms.

\item \textbf{Innovative Tract-Specific
Generation:} We train Tract-RLFormer to generate the tract of interest, utilizing our developed Mask Refinement Module (MRM) to generate tracking masks for the target tract, bypassing segmentation overhead.

\item \textbf{Generalization: }Through extensive testing on diverse datasets (TractoInferno, HCP, ISMRM2015), we demonstrate our network's superior performance and generalization capabilities across different neuroimaging contexts.
\end{enumerate}

\section{Related Work} 

Research in fiber tractography has transitioned from traditional deterministic and probabilistic methods to machine learning and deep reinforcement learning. Supervised learning and the exploratory dynamics of deep reinforcement learning unlock several possibilities for accurately mapping the brain's connectivity. Below, we review some recent works in these paradigms.

\textbf{Supervised Machine Learning based Algorithms: } In several studies, machine learning techniques have been explored to enhance fiber tractography with promising results. Notably, \cite{neher2017fiber,neher2015machine} utilized a Random Forest classifier in a supervised learning setting to identify $25$ distinct fiber bundles, leveraging data from the ISMRM2015 dataset \cite{maier2017challenge}. The effectiveness of their approach was assessed using the Tractometer tool, demonstrating the classifier's ability to accurately distinguish between different fiber pathways. Building on this foundation, subsequent research shifted focus towards regression-based methods for fiber tracking. \cite{poulin2017learn} suggested to employ a Gated-Recurrent Unit (GRU) model to predict new tracking steps from diffusion signal resampled to $100$ directions. This method advances the field, moving beyond traditional classification techniques to offer a more nuanced understanding of fiber tract development. Advancing the understanding of deep learning's potential for tractography, \cite{benou2019deeptract} applied both deterministic and probabilistic approaches to the task. In \cite{wegmayr2018data}, authors introduced an innovative method known as iFOD3, utilizing a feed-forward neural network to analyze raw, resampled signals. This approach considers the spatial context of streamlines, incorporating seed points located at the interface between white and gray matter—a notable departure from conventional methods that focus solely on white matter. This broader perspective on seed point placement contributed to the method's enhanced performance. In a subsequent development in \cite{wegmayr2021entrack}, authors presented a probabilistic machine learning model that outputs Fischer-von-Mises distributions rather than deterministic paths. This approach marked improvement over previous techniques, offering a more accurate and effective means of mapping the intricate networks of brain fiber tracts. These advancements underscore the rapidly evolving landscape of fiber tractography, highlighting the critical role of machine learning and deep learning in pushing the boundaries of neuroimaging research.

\textbf{Reinforcement Learning based Algorithms: }Contrary to the supervised training common in machine and deep learning approaches (poses challenges due to the difficulty of generating large scale ground truth data), the authors explored reinforcement learning (RL)-based approach for fiber tractography in \cite{theberge2021track}. In this approach, tractography is conducted similar to classical methods, wherein a reward function is employed by a learning model to generate streamlines based on local fiber orientation. Unlike the supervised paradigm for tractography, the RL-based model does not utilize reference streamlines while training. In \cite{theberge2021track}, the Twin-Delayed Deep-Deterministic Policy Gradient (TD3) algorithm \cite{fujimoto2018addressing} and the Soft Actor-Critic (SAC) algorithm were employed for RL-based fiber tractography to reduce false positives and enhance model generalization. In \cite{theberge2024matters}, the authors further examined different aspects of the RL framework, such as algorithm choice, seeding strategies, state representation, and reward functions, paving the way for advancements in this domain.

\section{Proposed Methodology}
\label{sec:method}
In this section, we begin by discussing the data and its preprocessing, followed by a systematic presentation of our proposal.
We utilize three public diffusion MRI datasets (Table \ref{tab:datasets}).
\begingroup
\begin{table}[]
\caption{Description of the three public DWI Datasets}
\begin{tabular}{ p{0.2\textwidth}p{0.13\textwidth}p{0.35\textwidth}p{0.32\textwidth} }
 \hline
 \textbf{Dataset}& \textbf{Subjects} &\textbf{DWI data}&\textbf{Distortion Corrections}\\
 \hline
 \textbf{TractoInferno \cite{poulin2022tractoinferno}}   & 284    &b=1000 $s/mm^{2}$; resolution=1mm isometric&   N4 bias field; eddy-current; head-motion \\

 \hline
 \textbf{HCP \cite{van2012human}}&   1200  &b=1000/2000/3000 $s/mm^{2}$; 270 directions; resolution= 1.25mm isometric  &EPI; eddy-current; subject-motion \\

 \hline
 \textbf{ISMRM \cite{maier2017challenge}} &1 & b=1000 $s/mm^{2}$; 32 directions; resolution=2mm isometric &  eddy currents; head motion \textit{(by our preprocessing)} \\

 \hline
\end{tabular}
\label{tab:datasets}
\vspace{-0.5cm}
\end{table}
\endgroup 
These datasets include a series of diffusion weighted images (DWI) that capture the diffusion of water molecules in tissue. Each voxel in a DWI contains information about the magnitude and direction of water diffusion, reflecting the underlying tissue micro-structure.

\textbf{Diffusion MRI Pre-processing: }We process DWI data to extract crucial information, including Spherical Harmonics Coefficients (SHC), Fiber Orientation Distribution Functions (fODF), and fiber peaks. Initially, the DWI data is projected into an $8^{th}$ order spherical harmonics basis, yielding 45 SHC volumes. The fODF, representing the distribution of fiber orientations within each voxel, is then computed, providing essential local information regarding streamline orientation. 
Subsequently, using the fODF, local fiber directions (peaks) are computed, which are used to define the reward function for training networks in the RL framework, as elaborated in Section \ref{subsec:policy-gen}.

\begin{figure}[hbt]
    \centering
    \vspace{-0.35cm}
    \includegraphics[width=12cm]{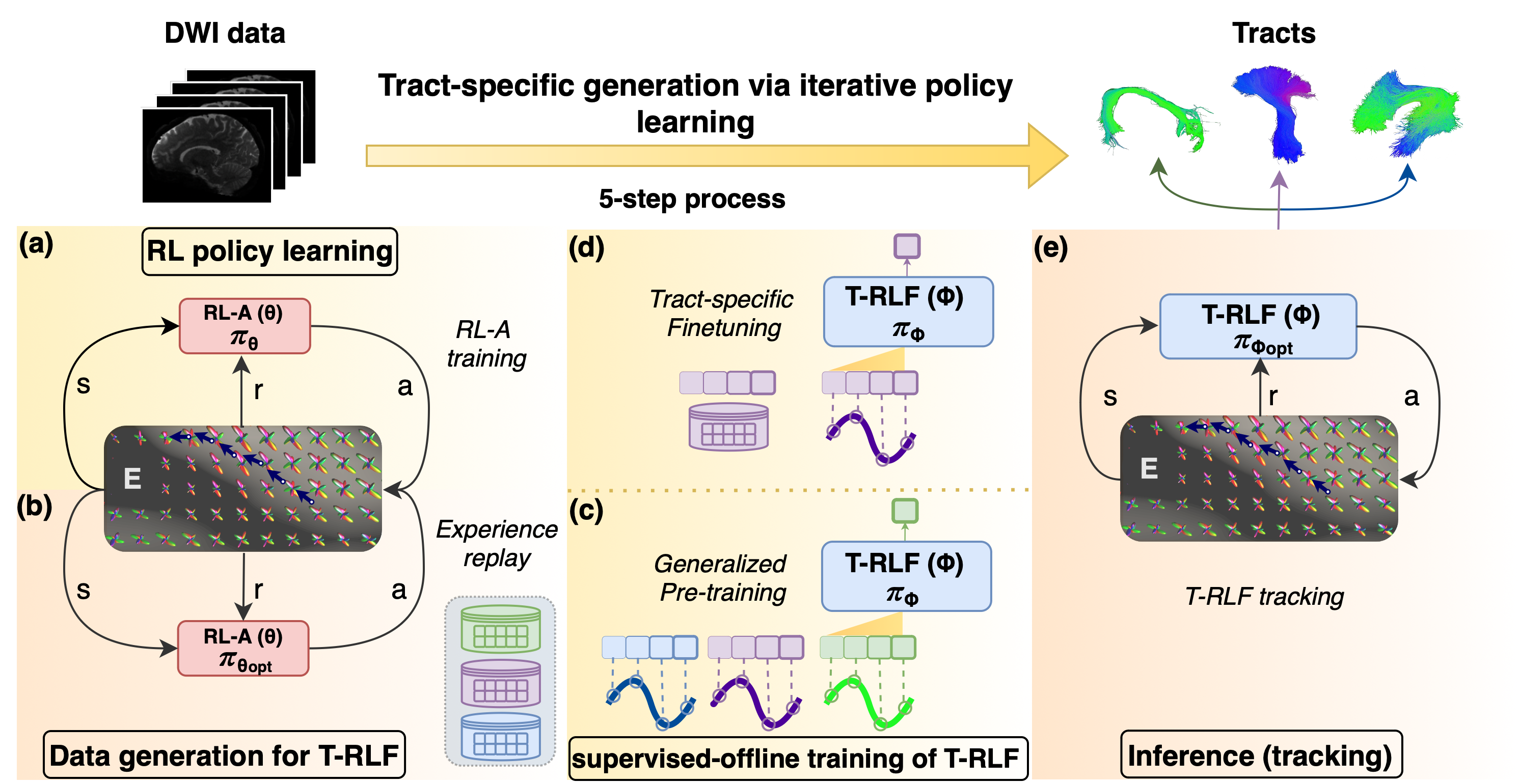}
    \caption{Overview of the proposed Iterative Policy Learning for Tract-Specific Generation using DWI data. (a) An RL agent ($\pi_{\theta}$) interacts with the environment (E) to learn an optimal \textbf{level-1 policy ($\pi_{\theta opt}$)}. (b) This policy is used to generate tract-specific roll-outs, denoted as 'experience replay'. (c) and (d) illustrate the offline, auto-regressive training of the proposed Tract-RLFormer $\phi$, referred to as T-RLF, over these roll-outs. In (c), T-RLF undergoes general pre-training, while in (d) it is fine-tuned to learn an optimal tract-specific policy ($\pi_{\phi opt}$). (e) shows the testing phase, where T-RLF, which has learned the new \textbf{level-2 policy ($\pi_{\phi_{opt}}$)}, performs tracking in environment $E$ to produce the desired tract. Training and tracking steps are shown in yellow and orange backgrounds, respectively.}
\label{fig:fig1}
    \vspace{-0.5cm}
\end{figure}

Moreover, in traditional tractography methods, white matter masks are typically derived from DWI data to perform whole-brain tractography, followed by segmentation of specific tracts. In contrast, we generate tailored masks for each tract, as detailed in Section \ref{subsec:mask-gen}. Our models are trained and tested within these masks, allowing for precise and efficient tract-specific analysis.\\\\
We propose an \textit{iterative policy learning framework for tract-specific generation}, delineated as a five-step process (see Fig. \ref{fig:fig1}).
In this framework, we start by training an RL agent (TD3) to learn a policy by exploration (within the tracking mask) to generate a tract of interest. We call it as level-1 policy.
Using this initial policy, the agent interacts with the (tracking) environment by taking actions (tracking steps). The agent’s experience (policy rollouts) is collected and sampled to train a refined version of the policy, by our T-RLF model, which learns in a data-driven manner through general pre-training and tract-specific fine-tuning. 
Our study focuses on seven principal white matter (WM) tracts: Corpus Callosum (CC),  left and right Pyramidal (PYT), Arcuate Fasciculus (AF), and Cingulum (CG) Tracts. The selection of these seven tracts is based on their clinical significance and frequent analysis as suggested in \cite{rheault2019bundle,theberge2024matters}.
To conduct such tract-specific training and generation, we first compute a tracking region of interest (mask) tailored for each tract using our Mask Refinement Module (MRM), described in \ref{subsec:mask-gen}. 
Following this, we proceed with the five sequential steps depicted in Fig. 1, detailed in subsequent subsections of the methodology.

\subsection{Mask Refinement Module (MRM)} \label{subsec:mask-gen}

We combine reference tracts from 2 Atlases, namely HCP842 \cite{yeh2018population}, and RecobundlesX \cite{francois_rheault_2023_7950602} to develop a fiber template for each of the seven tract classes. To obtain the mask of a given tract for any subject, the template fibers of the tract are aligned to the subject's brain space \cite{avants2009advanced}, creating an initial mask which is then dilated by 5 millimeters to get an augmented region of interest (ROI). This ROI is further refined by our Mask Refinement Module (MRM), which produces a tracking mask for a specific tract utilizing the fiber orientation information of the given subject. 
It consists of a fully connected neural network (FCNN) that refines the augumented ROI to obtain an estimate of the ground-truth mask for a given subject.
The process starts with a larger mask and progressively refines it by eliminating its voxels based on the Spherical Harmonics Coefficients (SHC) in the local neighborhood. The input for each voxel is the SHC (45 per voxel) of the voxel itself and its six immediate neighbors, concatenated with the expanded mask values, resulting in an input size of 322 (7 * 46).

The neural \textbf{network architecture} comprises three hidden layers with 512, 256, and 128 neurons, respectively. Each layer employs a ReLU activation function and is followed by batch normalization and a dropout layer (0.5). The output layer uses a sigmoid activation function, which determines the probability of retaining each voxel in the refined mask. Voxels with an output probability greater than 0.5 are kept in the predicted mask, while those with lower probabilities are eliminated. \textbf{Training} is performed voxel-wise, using Binary Cross Entropy as the loss function to compare the predicted mask value with the ground truth for each voxel. The model was trained with 50 subjects randomly selected from the TractoInferno dataset over 100 epochs. The resulting mask is then dilated by 1mm to produce the final refined tracking mask for the given subject.

\subsection{RL Policy Learning}
\label{subsec:policy-gen}
We learn a Level-1 policy by training a reinforcement learning (RL) agent ($\pi_{\theta}$) to perform fiber tracking. The RL agent learns the policy through exploration within the tracking environment (E) (see Fig. \ref{fig:fig1} (a)).

\textbf{Environment Details: }
Adopting the RL framework from \cite{theberge2021track}, we train an RL agent within the 3D diffusion MRI voxel space. The training process starts from seed voxels chosen within a 3D tract-specific mask (M), obtained from MRM. At any given voxel, the environment presents state ($s_t$) to the agent and rewards the agent's actions based on their alignment with the fODF peak, aiding in the learning of the optimized policy $\pi_{\theta opt}$. The tracking continues until the streamline exits the mask (M), surpasses a maximum length (l), or deviates significantly (>60°) from the previous tracking direction.

The \textbf{state ($s_t$)} is defined by 45 spherical harmonic (SH) coefficients and tracking mask (M) values from the current and six neighboring voxels, along with the four previous tracking directions, amounting to 334 dimensions (7×(45+1)+ 3×4). The predicted \textbf{action ($a_t$)} is a 3D vector representing tracking/fiber direction. The action space of the environment is continuous, allowing the agent to explore a wide range of potential fiber directions, with values in the range [-1, 1].
The \textbf{reward ($r_t$)} at time-step $t$ is given by the absolute dot product between the agent's predicted action ($a_t$) and the closest fODF peak ($p_i$), weighted by the dot product of the action ($a_t$) with the agent's previous tracking step ($\vec{u}_{t-1}$) (defined below).

\begin{equation}
r_t = \left| \max_{\vec{p}_i} \left( \vec{p}_i \cdot \vec{a}_t \right)\right| \times \left( \vec{a}_t \cdot \vec{u}_{t-1} \right)
\end{equation}

\textbf{Training details: }
During the agent's exploration phase, the transitions ($s,a,r,s'$) are recorded in a replay buffer for batch-wise policy optimization.
We utilize the TD3 algorithm to train 7 tract-specific agents. It has an Actor and two Critic networks (along with their time delayed target networks).

The actor and critic networks are both fully-connected neural networks with two ReLU activated hidden layers of 1024 neurons each. The actor has a 334 dimensional input layer and 3-neuron tanh activated output layer, while the critic has a 337 dimensional input layer and a single neuron tanh output layer (similar to \cite{theberge2021track}). Each tract-specific RL agent is trained on five subjects from the TractoInferno dataset (1030, 1079, 1119, 1180, and 1198), for 50 batches (4096 episodes each) per subject, hence a total of 1,024,000 (250*4096) episodes. We train the TD3 agent in 5 different instances of the environment (E) specified by each subject's distinct diffusion data, fODF peaks, and tracking mask. Training is conducted at 7 seeds per voxel and a step-size of 0.375mm, with fiber lengths between 20mm and 200mm. Maximum possible episode length is set to 530 (200/0.375). Other hyper-parameters include: learning rate: 8.56e-06, Discount factor ($\gamma$): 0.776, and Exploration noise ($\sigma_{train}$): 0.334.

\subsection{T-RLF: Policy Refinement}
\label{subsec:prp}

This subsection involves the training steps of our T-RLF model. We train a GPT-based network, to learn a refined, $level-2$ policy ($\pi_{\phi opt}$) for tract-specific fiber generation. It is trained on the policy rollouts of the $level-1$ TD3 policy ($\pi_{\theta opt}$) (ref: Section 3.2) to interpret and generate fiber data within the agent's experience space. This is accomplished through a two-stage process: (a) Initially, the Tract-RLFormer undergoes a generic, tract-agnostic \textbf{pre-training.} (b) This is followed by \textbf{fine-tuning} for the downstream task of tract-specific generation. Together, these constitute the next three steps (out of five), namely training data generation (Fig. \ref{fig:fig1}(b)) and the two-stage training process (Fig. \ref{fig:fig1}(c, d)) of T-RLF. Each component of the training framework is discussed in detail below.

\begin{figure}[hbt]
\vspace{-0.3cm}
    \centering
    \includegraphics[width=10cm]{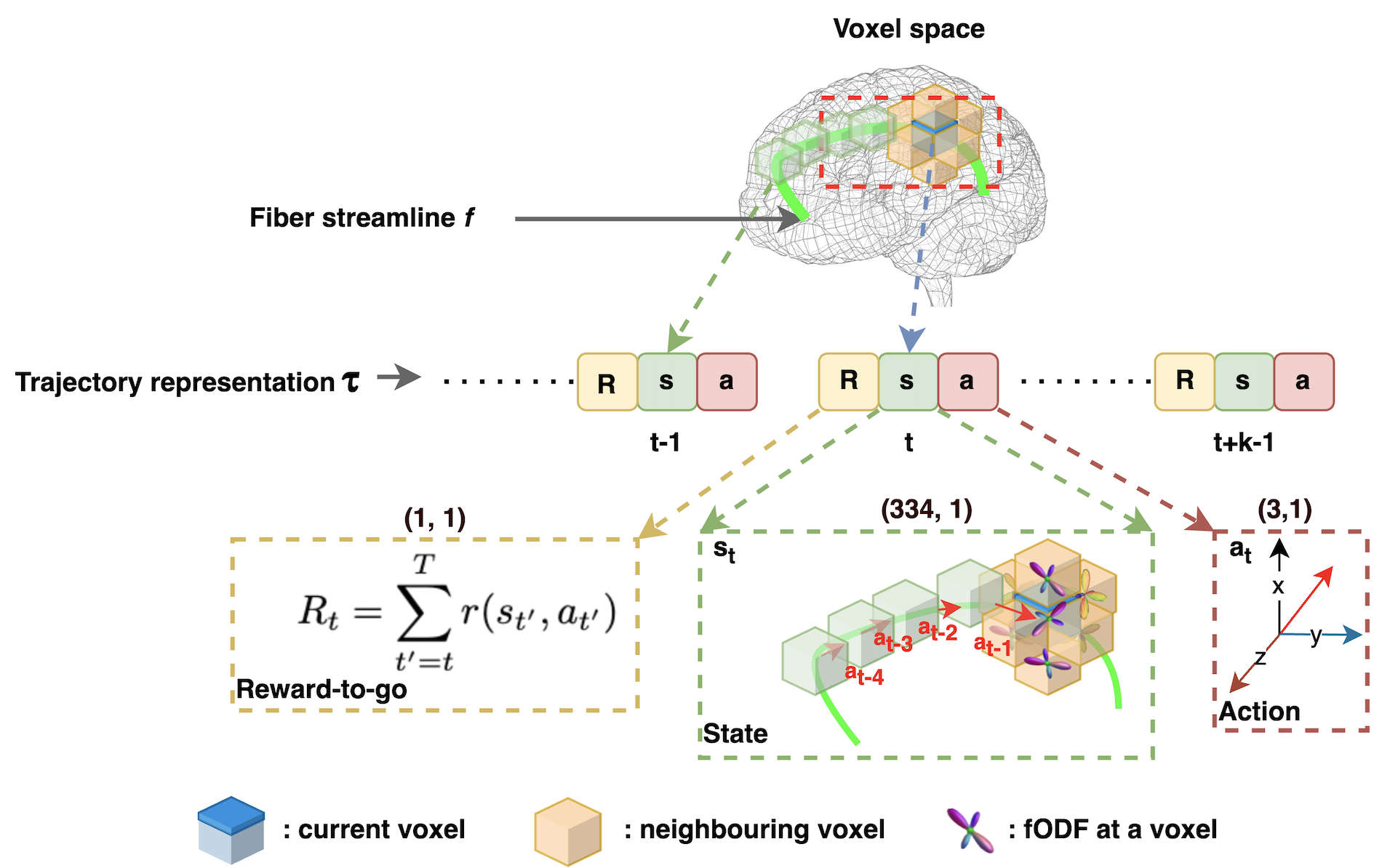}
    \vspace{-0.25cm}
    \caption{ Data Representation for T-RLF: Tract specific policy refinement using a trajectory-based approach in an RL agent's experience space. The figure illustrates a $k$ length fiber streamline $f$ in human brain voxel space, represented as a trajectory $\tau = (R_0, s_0, a_0, R_1, s_1, a_1, ....., R_k, s_k, a_k)$. Each point in the streamline corresponds to a state, action, and return-to-go tuple at a time-step $t$.}

    \label{fig:data-input}
    \vspace{-0.5cm}
\end{figure}

Unlike prior methods that generate fiber points by training on diffusion information along ground truth fiber streamlines, our network, T-RLF, learns from the sequence of state-action-reward (s, a, r) tuples (policy roll-outs) of a trained RL agent (Fig. \ref{fig:data-input}). T-RLF is trained on trajectories derived from the policy roll-outs of seven tract-specific TD3 agents. Each trajectory is represented as $\vec{\tau = (R_{0}, s_0, a_0, R_{1}, ..., R_{T}, s_T, a_T)}$, where $\vec{R_{t}}$ is the scalar sum of rewards from time-step $t$ to the episode's end, $\vec{s_{t}}$ is a 334-dimensional state vector, and $\vec{a_{t}}$ is a 3-dimensional action (see Section \ref{subsec:policy-gen}).

\textbf{Training Data Generation: }
To generate training data trajectories, we initiate tracking for the 7 trained TD3 agents (see Section \ref{subsec:policy-gen}) on 5 training subjects from the TractoInferno dataset. For each of the 7 tracts, we save all tracking episodes (until termination) as $(R, s, a)$ sequences, called trajectories. Tracking is conducted for all 5 subjects within their tract-specific masks using 7 seeds per voxel, resulting in a total of $\sum_{s=1}^{5} 7 \times {n_{v_{s,i}}}$ tract-specific trajectories for the $i^{th}$ tract, where ${n_{v_{s,i}}}$ is the number of voxels in the $i^{th}$ tract's mask for the $s^{th}$ subject. From these, 50,000 trajectories are selected per tract, with 10,000 from each subject. Half of these (5,000) are the longest trajectories for that subject’s $i^{th}$ tract, while the other half represent the streamline variability of the tract. This yields a tract-specific dataset $\tau_{i}$ for each tract $i$ used for model fine-tuning for downstream tasks.
From the 350,000 (7 x 50,000) trajectories of the seven tract-specific datasets ($\tau_{i}$ for $i=1\ to\ 7$), a total of 150,000 trajectories are selected. Half (75,000) of these are the longest trajectories, and the other half are randomly selected, resulting in a mixed tract dataset $\tau_{mix}$ used for generic tract-agnostic pre-training.

It should be noted that ${n_{v_{s,i}}}$ varies with the tract and subject. ${n_{v_{s,i}}}$ is the total number of voxels within the tracking mask for tract $i$ $(M_i)$ when aligned to the space of subject $s$.
Moreover, the minimum and maximum length of trajectories in the $\tau_{mix}$ dataset (representative of all tracts) are 48 and 292 respectively. It is later used to determine the training parameter of GPT model.

\begin{figure}[hbt] 
\vspace{-0.5cm}
    \centering
    \includegraphics[width=10cm]{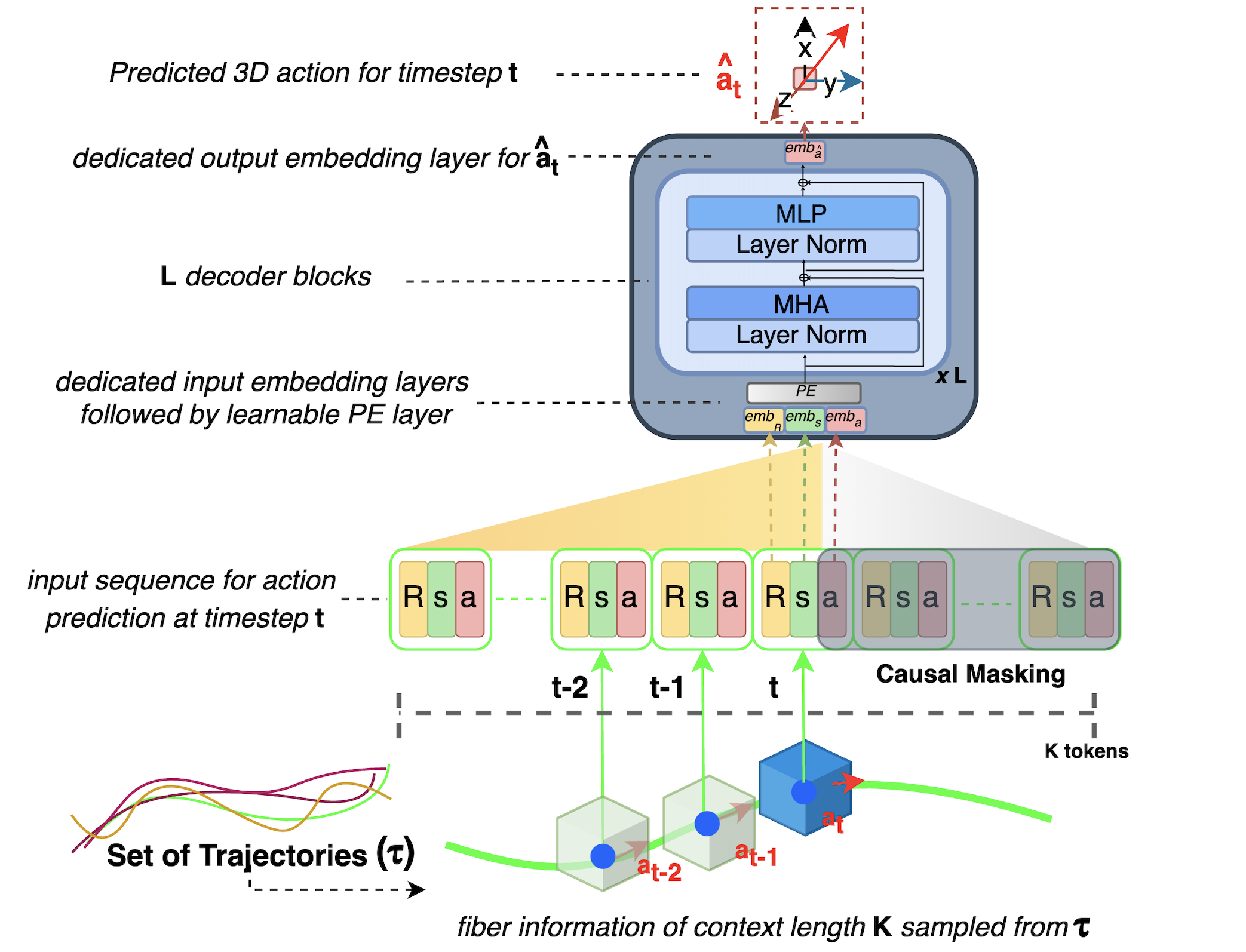}
    \vspace{-0.25cm}
    \caption{Data Driven Policy Learning: Visual representation  of training Tract-RLFormer for action prediction at time-step $t$, using context information from $K$ length fiber (Section \ref{subsubsec:train-det}). The input sequence tuples <$R$, $s$, $a$> are causally masked from $a_t$ onwards and processed through embedding layers $emb_R$, $emb_s$, and $emb_a$, with a learnable positional encoding layer ($PE$). Embeddings are processed by $L$ decoder blocks ($L=3$ for pre-training, $L=4$ for fine-tuning), incorporating Multi-Head Attention (MHA) and Multi-Layer Perceptron (MLP), to generate predicted action $\hat{a_t}$.   
}
    \label{fig:training}
    \vspace{-0.7cm}
\end{figure}

\textbf{Model Architecture: }Tract-RLFormer adopts the GPT architecture (as shown in Fig. \ref{fig:training})  to model trajectories autoregressively \cite{chen2021decision}. The network consists of 4 decoder layers with 1 attention head each ($n\_heads$=1), a context length ($K$) of 40, an embedding dimension ($d)$ of 128, ReLU activation functions, and a dropout rate of 0.1.
These parameters were selected after thorough experimentation
presented in section \ref{subsec:ablation}.
It begins with a dedicated embedding layer of 128 dimensions (as shown in Fig. \ref{fig:training}) for each component of the trajectory: state (s), action (a), and return-to-go (R). Subsequently, a trainable positional encoding layer processes the timestep sequence (of $max\_ep\_len$) as input, generating positional/timestep embeddings of dimensionality $d$ = 128, where each timestep (t) has 3 tokens <$R_t$, $s_t$, $a_t$>.
The maximum possible episode length ($max\_ep\_len$) controls length of episode. It is set to 530 because the maximum length of a fiber is 200mm, equivalent to 530 steps for a TractoInferno subject (as 1 step corresponds to 0.375 mm; refer \ref{subsec:infer}).
If an episode exceeds 530 timesteps, it is truncated to this length.
Embeddings for each component of the trajectory (state, action, return-to-go) are then combined and fed into the decoder layers. We utilize four decoder blocks, where each block includes a multi-headed self-attention mechanism followed by position-wise feed-forward networks. After processing through the decoder blocks, the output is passed through an output embedding layer, from which we obtain the predicted action of dimension (3, 1).

\textbf{Training Details: }
\label{subsubsec:train-det}
The proposed T-RLF model is trained to generate an optimal level-2 policy, ($\pi_{\phi opt}$), specifically tailored for tract-specific generation. It undergoes a two-stage training process, starting with general pre-training on mixed tract dataset ($\tau_{mix}$), followed by tract-specific fine-tuning on tract-specific dataset ($\tau_i$). 
The first three decoder layers are pre-trained over $0.15$ million mixed trajectories (taken from $\tau_{mix}$), containing a total of 30 million transitions for 30 iterations. Later the $4^{th}$ decoder layer is fine-tuned on the tract-specific trajectories buffer for 10 additional iterations. In each iteration, the model undergoes 10,000 training steps, each processing a $batch\_size$= 128 number of K-length trajectories. 
A batch of $128$ tokens of $<R_{t}, s, a>$ are sampled from training data ($\tau$) and stacked for a context length ($K = 40$) and fed as an input to the T-RLF. It passes through an embedding layer with 128 dimensions, and positional encoding is added, resulting in a (128 x 120 x 128) matrix and is processed by the 4 decoder layers with causal masking (Fig. \ref{fig:training}). The decoder output is mapped through an output embedding layer to predict the action. 
Unlike the TD3 agent, T-RLF does not interact with the environment during its training process. Instead, it is trained entirely in an offline mode using only trajectory datasets ($\tau$'s).
For the context length K, a 5-step loss (accounting for current and 2 steps in both forward and backward directions) is computed, aggregating the angular difference between predicted and actual action at each time-step.
\begin{equation}
    L = \sum_{t=2}^{K-2} \left( \sum_{i=-2}^{2} \cos^{-1} \left( \textbf{a}_{t+i} \cdot \hat{\textbf{a}}_{t+i} \right) \right)
\end{equation}

The learning of weights for $\pi_{\phi opt}$ is facilitated by this 5-step loss function, in order to generate more effective and robust actions. Here, $\vec{a}_{t+i}$ and $\vec{\hat{a}}_{t+i}$ are the true and predicted actions at $(t+i)^{th}$ timestep respectively.

Similar to \cite{chen2021decision}, T-RLF training is conditioned to generate action ($a_{t}$) using return ($R_{t}$) at each timestep.
During inference, $R_{t}$ is initialized to an expert return value or the longest trajectory return. In our case, the longest trajectory length is 292, and since the maximum possible reward at each timestep is 1, we initialize $R_{t}$ to 300 ($\thicksim$1x expert return). This was experimentally verified among various values: 100, 200, 300, 500, and 600. For model training, we employed the AdamW optimizer, set with a learning rate of 1e-4 and a weight decay of 1e-4.

\subsection{T-RLF: Inference} \label{subsec:infer}
The final step in our fiber tract generation method involves using the trained T-RLF models to perform tracking, followed by cleaning the resulting tracts. Having learnt the refined policy ($\pi_{\phi opt}$), T-RLF can function autonomously as a generic substitute for TD3 agent. \textbf{Consequently, it can independently perform fiber streamline generation in the same environment ($E$)} as detailed in section \ref{subsec:policy-gen}, without relying on the original TD3 agent.
Fiber generation (tracking) is executed within tract-specific masks obtained from MRM and is initialised with 7 seeds per voxel, and $R_t$ is set to $R_0=300$. Tracking step size is (empirically  selected) and is dataset-specific, \textbf{$0.375mm$} for the TractoInferno, \textbf{$0.468mm$} for HCP, and \textbf{$0.75mm$} for the ISMRM dataset. At each step, the return-to-go ($R_t$) is reduced by the achieved reward and predicted action($a_t$), new state ($s_t'$), and $R_t$ are appended to the context window to serve as input for the next prediction. This auto-regressive process by Tract-RLFormer generates the fiber tract of interest. Finally, the tracts undergo a \textbf{Cleaning procedure} using a fast streamline search (FSS) \cite{st2022fast} to eliminate any extraneous fibers, by comparing the predicted tract with the atlas reference tracts (representing general anatomical structure).
Our tracts are confined to masks generated by the MRM module, tailored to each subject's fiber orientation. This approach ensures that tract generation remains confined to regions proximate to the actual neural fibers of the subject, thus mitigating the risk of false positives. Consequently, we can perform a high radius search using FSS, without incurring a major risk of high overreach. This high radius search ensures that accurate fibers are not discarded based on minor deviations from atlas tracts.

\textbf{Performance Parameters: }In order to evaluate the quality of our generation, the Ground Truth tract is aligned to Montreal Neurological Institute (MNI) space using Advanced Normalization Tools (ANTs) \cite{avants2009advanced}, facilitating comparison with our cleaned tracts that are already in MNI space. The Dice ($D$), Overlap ($OvL$), and Overreach ($OvR$) scores (similar to \cite{theberge2021track,theberge2024matters}) are then computed against the ground truth tract and are reported in Section \ref{sec:res-and-discuss}. 
The Dice score assesses both the accurate coverage and the minimization of extraneous extensions beyond the ground truth area, where values near 1 signify a high similarity level.
Overlap measures the intersection of the generated tract with the ground truth, while Overreach indicates how much the generated tract exceeds the ground truth, with lower scores suggesting greater precision.

\section{Results \& Discussion} \label{sec:res-and-discuss}
In this section, we present the outcomes of our evaluation of tract-specific T-RLF models under various experimental setups, including comparative analysis, generalization performance, and an ablation study. 
We trained TD3 and T-RLF models, on eight tracts— seven principal white matter tracts (refer Section \ref{sec:method}) and OR tract (for analysis in \ref{subsec:comp-analysis}) using five \textbf{train subjects} (id: 1030, 1079, 1119, 1180, and 1198) of the TractoInferno dataset and reported their performance on various test subjects across different datasets in subsequent subsections.
Additionally, we assess their effectiveness relative to supervised approaches and traditional tractography methods that do not incorporate learning.

\subsection{Comparative Analysis}
\label{subsec:comp-analysis}
This section provides a comparative analysis of our model, T-RLF, against supervised learning, traditional tractography, and state-of-the-art (SOTA) reinforcement learning (RL) algorithms, using Dice scores to evaluate performance across three major white matter bundles: $PYT, OR,$ and $CC$. As presented in Table \ref{table:comparetive}, all methods are tested on \textbf{subject 1006} from the TractoInferno dataset (similar to \cite{theberge2021track,theberge2024matters} for fair comparison). For the first and third tabular subparts of Table \ref{table:comparetive}, the models are trained on ISMRM data. The second subpart does not involve training (classical methods). These 3 subparts are assessed using whole-brain tractography and segmentation \cite{poulin2022tractoinferno}\cite{theberge2024matters}. Additionally, the last subpart details the performance of our T-RLF and the TD3 model, where T-RLF was specifically trained on trajectories derived from the TD3 agent.

\begin{table}[hbt]
\caption{Comparison of mean Dice scores for the OR, PYT, and CC tracts for subject 1006 from TractoInferno dataset. Supervised learning scores are from \cite{poulin2022tractoinferno}; RL-based scores, with std. dev., are from \cite{theberge2024matters}. The last 2 rows includes scores for T-RLF and TD3, evaluated using our tract-specific approach. The highest and second highest scores are highlighted in green and red, respectively. `*' denotes tract-specific setting for methods.}
\begin{tabular}{ |p{4cm}|p{2.5cm}|p{2.5cm}|p{2.5cm}|  }
 \hline
\textbf{Algorithm}& \textbf{OR} &\textbf{PYT}&\textbf{CC}\\\cline{1-4}
  DET-SE   & 0.569    &0.665 &   0.658\\
  DET-Cosine&   0.598  & 0.708   &0.646\\
Prob-Sphere &0.599 & 0.695&  0.648\\
  Prob-Gaussian    &0.542 & 0.723&  0.668\\
  Prob-Mixture&   0.436  &  0.522& 0.614\\\cline{1-4}
 \hline
  DET&   0.516  &  0.475& 0.345\\
  PROB&   0.549  &  0.740& 0.590\\
  PFT   & $0.644 \pm 0.136 $  &$0.753 \pm 0.010 $ &   $\colorbox{SpringGreen}{0.827} \pm 0.008 $\\ \cline{1-4}
  VPG& $0.369 \pm 0.135 $  & $0.434 \pm 0.128 $  & $0.428 \pm 0.182 $ \\
A2C  & $0.225 \pm 0.108 $ & $0.323 \pm 0.082 $&  $0.222 \pm 0.025 $\\
ACKTR  & $0.397 \pm 0.171 $ & $0.559 \pm 0.028 $&  $0.584 \pm 0.054 $\\
TRPO  & $0.330 \pm 0.154 $ & $0.498 \pm 0.062 $&  $0.594 \pm 0.048 $\\
PPO  &$0.440 \pm 0.187 $ & $0.619 \pm 0.042 $&  $0.650 \pm 0.028 $\\
DDPG &$0.612 \pm 0.063 $ & $0.630 \pm 0.045 $& $0.731 \pm 0.006 $ \\
TD3  & $0.555 \pm 0.097 $ & $0.603 \pm 0.045 $&  $0.688 \pm 0.035 $\\
  SAC    & $0.598 \pm 0.098 $ & $0.658 \pm 0.028 $&  $\colorbox{IndianRed1}{0.753} \pm 0.010 $\\
  SAC Auto& $0.608 \pm 0.088 $ & $0.655 \pm 0.032 $ & $0.747 \pm 0.019 $\\\cline{1-4}
 \hline
DET$^*$ & 0.648 & 0.752 & 0.713 \\
PROB$^*$ & \colorbox{IndianRed1}{0.652} & 0.765 & 0.731 \\
TD3$^*$ & 0.644 & \colorbox{IndianRed1}{0.764} & 0.720 \\
  T-RLF (Ours)   & \colorbox{SpringGreen}{0.673}    &\colorbox{SpringGreen}{0.772} &   0.738\\
 \hline
\end{tabular}
\label{table:comparetive}
\end{table}

In Table \ref{table:comparetive}, our framework outperforms the state-of-the-art method (PFT) for PYT and OR tracts, demonstrating its robustness in tract-specific tractography. Additionally, T-RLF shows comparable performance to state-of-the-art RL algorithms for CC tract.

Furthermore, the TD3 agent demonstrates markedly improved performance within our tract-specific generation framework. 
Testing of tract-specific TD3 on the ISMRM or HCP datasets cannot be conducted due to the absence of the evaluated tracts in these datasets.
However, the enhancement in TD3's performance in our tract-specific setting can be attributed to the training approach rather than dataset consistency. This is evidenced by TD3’s comparable or superior performance on different tracts across the ISMRM and HCP datasets, as detailed further in Tables \ref{table:exp:cg_af}, \ref{table:exp:pyt_cc}.

Moreover, dice scores for DET and PROB improved for all tracts in the tract-specific setting, especially for CC, where DET increased by 106.67\% (0.345 to 0.713) and PROB by 23.89\% (0.590 to 0.731). The enhanced tracking performance of DET and PROB, despite not being trained, is indicative of the effectiveness of our tract-specific masks.
Also, in the whole-brain setting, there is a huge difference between DET (0.475) and PROB (0.740) scores on the PYT tract (Table \ref{table:comparetive}), whereas this gap is significantly smaller in the tract-specific setting (marked with `*'), where the tract-specific performance of DET$^*$ (0.752) and PROB$^*$ (0.765) align closely with each other and with the T-RLF and TD3 methods.
The consistency and stability observed for these classical methods are attributed to our tract-specific approach.

\subsection{Generalization Performance Evaluation}
\begin{figure}[hbt]
    \centering
    \includegraphics[width=12cm]{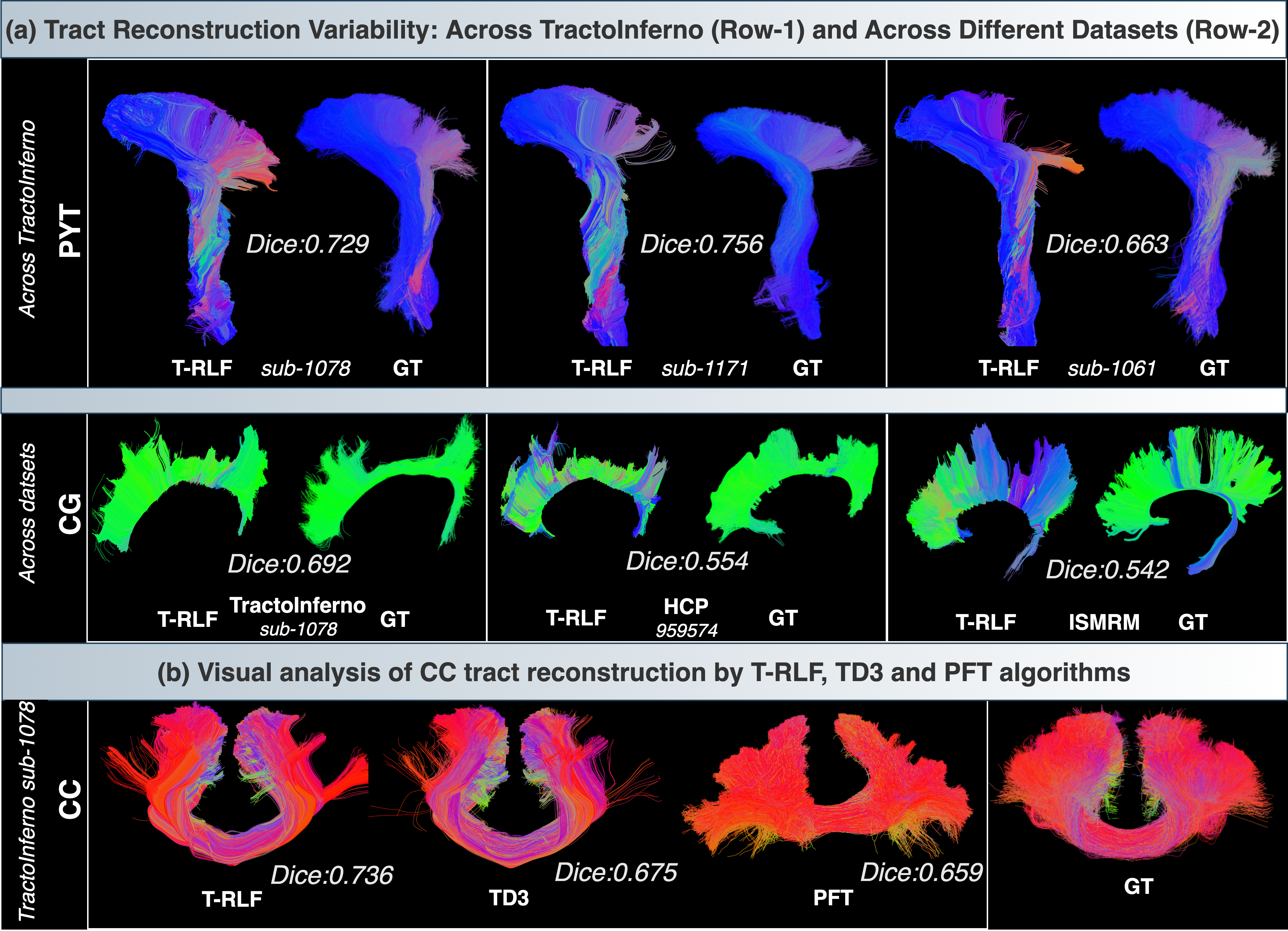}
    \vspace{-0.3cm}
    \caption{Visual comparison of reconstructed tracts illustrating (a): Intra-dataset variability, Inter-dataset variability, and (b): Variability across tracts reconstructed by different algorithms. The depicted tracts include the left PYT, CG, and a part of CC. The algorithms evaluated in bottom section of figure are T-RLF (ours), TD3, and PFT.}
    \label{fig:plot}
    \vspace{-0.3cm}
\end{figure}

In this section, we present the performance evaluation of our T-RLF model across three distinct datasets (Tables \ref{table:exp:cg_af}, \ref{table:exp:pyt_cc}), demonstrating its effectiveness and generalizability. The averaged results include analyses across five \textbf{test subjects} in the TractoInferno (TtoI) dataset (id: 1160, 1078, 1159, 1061, and 1171), four from the HCP dataset (id:  930449, 992774, 959574, and 987983), and one from the ISMRM dataset. A visual comparison across datasets and subjects is presented in Fig. \ref{fig:plot}(a). We also compare the performance of T-RLF with classical algorithms, which were employed using tract-specific masks, and the tract-specific TD3 agent, from which the training data for T-RLF was derived (Fig. \ref{fig:plot}(b)).

\begin{table}[hbt]
\centering
\caption{Performance metrics(in \%) for the CG and AF tracts, trained on the TractoInferno dataset and tested across multiple datasets to evaluate generalization. Tracking is performed using our proposed tract-specific generation method. A dash (`-') indicates the absence of ground-truth tracts in the corresponding dataset, precluding evaluation.
}
\begin{tabular}{ |p{1.18cm}|p{1cm}|p{0.7cm}p{0.7cm}p{0.7cm}|p{0.7cm}p{0.7cm}p{0.7cm}|p{0.7cm}p{0.7cm}p{0.7cm}|p{0.7cm}p{0.7cm}p{0.7cm}|  }
 \hline
    & &\multicolumn{6}{c|}{\textbf{Cingulum} (CG)}&\multicolumn{6}{c|}{\textbf{Arcuate Fasciculus}(AF)} \\\cline{3-14}
     & &\multicolumn{3}{c|}{Left} &\multicolumn{3}{c|}{Right} &\multicolumn{3}{c|}{Left} &\multicolumn{3}{c|}{Right} \\\cline{3-14}
\textbf{Dataset} &\textbf{Algo.}& Dice &OvL&OvR& Dice &OvL&OvR& Dice &OvL&OvR& Dice &OvL&OvR\\\cline{1-14}
  &T-RLF   & 53.3  & 42.5 & 16.6 & 45.6   &33.7  & 13.9 & 61.8  & 51.2 & 13.4 & 41.8   &27.9  & 5.40 \\
  &TD3 &  53.0  &42.3   &16.9  &45.2   &33.6  & 14.3 & 61.6  & 51.0 & 13.7 & 41.6   &27.7  & 5.60 \\\cline{2-14}
HCP  &DET & 55.2 & 46.4 & 21..5 & 52.6 & 41.7 & 16.3 & 62.8 & 52.5 & 14.0 & 43.9 & 30.0 & 6.6 \\
   &PROB    & 57.6 & 51.8 & 27.9 & 56.1 & 45.5 & 16.6  & 65.5 & 57.9 & 18.3 & 47.4& 33.7 & 8.6 \\
  &PFT  &  67.3 & 55.2 & 7.8 & 59.6  & 45.4  & 6.4  & 71.3  & 71.9 & 29.7 & 69.9  & 71.6 & 33.3 \\\cline{1-14}
  &T-RLF   & 61.0 & 56.8 & 28.6  &  56.5  &  52.6 & 34.9 & 52.7 & 45.1 &  27.8 &  39.5 & 36.5 & 49.8 \\
  &TD3 & 60.0  & 55.1  & 27.3 &  54.9  & 49.8  & 32.4 & 51.8  &  44.3 & 28.2  &  38.4   & 34.9 & 46.9 \\\cline{2-14}
 TtoI&DET & 61.2 & 58.8 & 32.3 & 58.2 & 54.7 & 33.7 & 54.6 &46.3 &24.7  & 45.4 & 41.7 & 46.9 \\
   &PROB    & 67.9 & 69.1 & 33.6 & 64.7 & 64.6 & 36.7  & 62.3 & 57.2 & 27.2 & 50.3 & 48.3 & 50.9 \\
  &PFT & 55.9  & 48.8 & 25.4 &  54.5   & 51.3  & 38.6  & 62.8  & 60.1 & 31.5 &  53.9 & 62.8 & 88.0 \\\cline{1-14}
  &T-RLF   & 54.2    & 46.6 & 25.5 & 52.8    & 44.1 &  23.1 &  - & - & - & - & - &- \\
  &TD3 &  53.1  & 44.8   & 23.7  &   51.2  & 41.6   & 21.1  &  - & - & - & - & - &-  \\\cline{2-14}
 ISMRM &DET & 57.5 & 51.9&  28.5 &57.7 & 52.4&  29.2 & - & - & - & - & - &- \\
   &PROB    & 61.1 & 59.3&  35.1 &64.0 & 65.4&  39.0  &  - & - & - & - & - &- \\
  &PFT & 55.4  &  49.4& 28.9 &   57.3  &  49.4& 22.9  &  - & - & - & - & - &- \\\cline{1-14}
 \hline
\end{tabular}
\label{table:exp:cg_af}
\vspace{-0.25cm}
\end{table}

In Table \ref{table:exp:cg_af}, we see that T-RLF model displays a notable generalization performance. Interestingly, the classical deterministic (DET) and probabilistic (PROB) methods exhibit slightly better performance than learnable methods in some cases (Tables \ref{table:exp:cg_af},\ref{table:exp:pyt_cc}).

\begin{table}[hbt]
\centering
\caption{Results are presented for the left and right parts of PYT and a segment of CC on the TractoInferno (TtoI) dataset. Tracking for all algorithms is conducted using our proposed tract-specific generation method.}
\vspace{-0.2cm}
\begin{tabular}{ |p{1.3cm}|p{1cm}|p{0.7cm}p{0.7cm}p{0.7cm}|p{0.7cm}p{0.7cm}p{0.7cm}|p{0.7cm}p{0.7cm}p{0.7cm}|  }
 \hline
    & &\multicolumn{6}{c|}{\textbf{Pyramidal Tract} (PYT)}&\multicolumn{3}{c|}{\textbf{Corpus Callosum}} \\\cline{3-8}
     & &\multicolumn{3}{c|}{Left} &\multicolumn{3}{c|}{Right} &\multicolumn{3}{c|}{(CC)} \\\cline{3-11}
\textbf{Dataset} &\textbf{Algo}.& Dice &OvL&OvR& Dice &OvL&OvR& Dice &OvL&OvR\\\cline{1-11}
  &T-RLF   & 70.3  & 64.1 & 17.2 & 70.1   &63.1  & 16.9 & 70.4  & 71.2 & 32.6  \\
  &TD3 &  69.4  &62.5   &15.9  &69.2   &61.4  & 15.8 & 68.1  & 64.8 & 26.1 \\\cline{2-11}
TtoI  &DET & 72.7 & 79.3 & 38.8 & 70.3 & 76.2 &40.7 & 70.1 & 72.6 & 35.8 \\
   &PROB    & 77.6 & 79.5 & 25.3 &74.8  & 72.5 & 21.3  & 72.6 & 76.4 & 36.7 \\
  &PFT & 66.2  & 55.7 & 12.4 & 65.9  & 57.8  & 17.4  & 54.9 & 51.2 & 36.1 \\\cline{1-11}
 \hline
\end{tabular}
\label{table:exp:pyt_cc}
\end{table}

As previously mentioned in section \ref{subsec:comp-analysis}, the consistency observed in Tables \ref{table:exp:cg_af},\ref{table:exp:pyt_cc} for the classical methods (DET and PROB) is due to our tract-specific approach. This improvement and stabilization may be attributed to the elimination of premature termination issues in narrow and deep WM regions, as described in \cite{girard2014towards}, facilitated by the refined spatial exploration enabled by MRM in our tract-specific approach.
It can be observed from Tables \ref{table:comparetive} and \ref{table:exp:pyt_cc}, that the performance of PFT declined in the tract-aware setting, dropping from 75\% to 66.2\% in the PYT and from 82\% to 55\% in the CC (refer Table \ref{table:exp:pyt_cc}).
This decline can be attributed to use of Continuous Map Criterion (CMC) as a stopping criterion for fiber tracking. The CMC terminates fiber tracking based on Partial Volume Estimate (PVE) maps, allowing tractography to continue until the streamline correctly stops in the gray matter. This approach may generate fibers beyond our tract-specific masks, leading to increased overreach (see Fig. \ref{fig:plot}(b)) and consequently lower Dice scores. Furthermore, fibers generated outside the tracking mask may be erroneous and subsequently filtered or cleaned via FSS, resulting in a lower OvL score.

\textbf{Summarization: }In summary, our results demonstrate that we surpass supervised methods (Table \ref{table:comparetive}). Additionally, we consistently outperform the TD3 model (Tables \ref{table:comparetive}- \ref{table:exp:pyt_cc}), which served as the basis for training T-RLF. Notably, our tract-specific setting not only improves TD3 performance but also the performance of classical methods like DET and PROB compared to the whole-brain setting. This suggests a promising new direction of data driven policy learning for tract specific fiber generation in limited ground truth scenarios that can naturally scale up effectively.

\subsection{Ablation Study}
\label{subsec:ablation}

We conducted an ablation study to determine the optimal configuration for our T-RLF model. The evaluation presented in Table \ref{table:dt-abl}, identified the best architecture with \textit{n\_heads}=1, \textit{K}=40, and an embedding dimension of \textit{d}=128. This study highlights the importance of a larger context in tractography, illustrating how a broader temporal receptive field can enhance the model's ability to generate accurate fiber tracts.

\vspace{-0.25cm}
\begin{table}[ht!]
\centering
\caption{Dice scores (in \%) averaged over 7 tracts of subject 1006 from TractoInferno dataset, at different values of T-RLF parameters: number of attention heads (\textit{n\_heads}), context length (\textit{K}), and embedding dimension (\textit{d}). Best score is in \textbf{bold}.}
\begin{tabular}{|c|c|c|c|c|c|c|}
\hline
\multicolumn{1}{|c|}{} & \multicolumn{2}{c|}{\textbf{$K = 20$}} & \multicolumn{2}{c|}{\textbf{$K = 30$}} & \multicolumn{2}{c|}{\textbf{$K = 40$}} \\
\cline{2-7}
\multicolumn{1}{|c|}{} & \textbf{ $d$=$128$ } & \textbf{ $d$=$512$ } & \textbf{ $d$=$128$ } & \textbf{ $d$=$512$ } & \textbf{ $d$=$128$ } & \textbf{ $d$=$512$ } \\
\hline
$n\_heads$ = 1 & 64.7 & 65.2 & 66.2 & 67.3 & \textbf{68.6} & 67.6 \\
$n\_heads$ = 2 & 63.4 & 66.3 & 66.2 & 67.5 & 68.0 & 68.2 \\
\hline
\end{tabular}
\vspace{-0.25cm}
\label{table:dt-abl}
\end{table}

We also examined the impact of two key components: Mask Refinement Module (MRM) discussed in section \ref{subsec:mask-gen}, and the tract-specific policy fine-tuning as detailed in section \ref{subsec:prp}.
Table \ref{table:abl:mrm} reports the results for the T-RLF network trained over TractoInferno dataset.  We have observed that initial tracking masks led to a significant overreach (OvR), extending beyond actual region of interest. This OvR was notably reduced after incorporating MRM, leading to improved Dice and overlap metrics across all tracts. Furthermore, fine-tuning the network specific to each tract allowed it to learn better and robust tract-specific diffusion characteristics, resulting in additional improvements in the performance metrics.

\begin{table}[hbt]
\centering
\caption{Average performance metrics (in \%) obtained using Tract-RLFormer highlight the impact of the MRM on test subjects from the TractoInferno dataset. The table also compares the performance of the pre-trained network with the fine-tuned network post MRM application, illustrating the effect of policy fine-tuning on the same dataset.}
\begin{tabular}{ |p{1cm}|p{1cm}p{1cm}p{1cm}|p{1cm}p{1cm}p{1cm}|p{1cm}p{1cm}p{1cm}|  }
 \hline
  \textbf{Tract}  &\multicolumn{3}{c|}{\textbf{Without MRM}} &\multicolumn{6}{c|}{\textbf{With MRM}} \\\cline{2-10}
  & & & & \multicolumn{3}{c|}{Pre-trained} & \multicolumn{3}{c|}{Fine-tuned} \\    \cline{5-10}
 & Dice &OvL&OvR& Dice&OvL&OvR & Dice&OvL&OvR\\
   \cline{5-10}
   \hline
   PYT& 44.1 & 30.4 & 5.6  &65.9 & 55.1 & 11.8 & 70.3 & 67.2 & 24.7 \\
   CG& 41.1  & 45.4 & 80.3 & 51.5 & 45.7 & 30.3 & 58.7 & 54.7 & 31.7 \\
   AF& 34.2  & 34.4 & 65.1  & 45.8& 39.5 & 36.1 & 46.1 & 40.8 & 38.8 \\
   CC& 58.9  & 59.0 & 41.9 & 66.2 & 59.9 & 20.8 & 70.4 & 71.2 & 32.6 \\
   \hline

\end{tabular}
\label{table:abl:mrm}
\end{table}

\section{Conclusion}
Tractography can be an essential tool in neuroimaging, enabling the detailed mapping of neural pathways crucial for both clinical and research applications. Our work significantly advances this field by introducing a data driven Tract-RLFormer framework which is a tract-specific, transformer-based network integrating supervised and reinforcement learning paradigms. A distinctive feature of our Tract-RLFormer is its ability to train within the reinforcement learning experience space, independent of ground truth fibers. The fine-tuning stage of our model focuses and refines its capabilities in generating the tracts of interest. This approach  demonstrates its excellent generalization performance across various datasets as well as scalability. Our data-driven approach has the potential to utilize data from any reinforcement learning agents trained in diverse neuroimaging environments. Moreover, our innovative tract-specific modeling approach simplifies the reconstruction process by directly generating the target tract, thus avoiding the complex and error-prone segmentation step.

\section{Acknowledgment}
This research was supported by SERB Core Research Grant Project No: CRG/ 2020/005492, IIT Mandi.

\bibliographystyle{splncs04}
\bibliography{references}
\end{document}